\documentclass{article}

\usepackage{arxiv}

\usepackage[utf8]{inputenc} 
\usepackage[T1]{fontenc}    
\usepackage{hyperref}       
\usepackage{url}            
\usepackage{booktabs}       
\usepackage{amsfonts}       
\usepackage{nicefrac}       
\usepackage{microtype}      
\usepackage{lipsum}
\usepackage{graphicx}
\usepackage{amsmath}
\usepackage{multirow}
\graphicspath{ {./images/} }

\title{N-DriverMotion: Driver motion learning and prediction using an event-based camera and directly trained spiking neural networks}

\author{
 Hyo Jong Chung \\
  College of Engineering and Applied Sciences\\
  Stony Brook University \\
  100 Nicolls Road, New York, USA \\
  \texttt{hyojong.chung@stonybrook.edu} \\
   \And
 Byungkon Kang \\
  Department of Computer Science\\
  The State University of New York Korea\\
  119-2 Songdo Munhwa-ro, Yeonsu-gu, Incheon, Korea \\
  \texttt{byungkon.kang@sunykorea.ac.kr} \\
  \And
 Yoonseok Yang \\
  Department of Computer Science\\
  The State University of New York Korea\\
  119-2 Songdo Munhwa-ro, Yeonsu-gu, Incheon, Korea \\
  \texttt{yoonseok.yang@sunykorea.ac.kr} \\
}

\begin{document}
\maketitle
\begin{abstract}
Driver motion recognition is a key factor in ensuring the safety of driving systems. This paper presents a novel system for learning and predicting driver motions, along with an event-based (720x720) dataset, N-DriverMotion, newly collected to train a neuromorphic vision system.
The system includes an event-based camera that generates a driver motion dataset representing spike inputs and efficient spiking neural networks (SNNs) that are effective in training and predicting the driver's gestures. The event dataset consists of 13 driver motion categories classified by direction (front, side), illumination (bright, moderate, dark), and participant.
A novel optimized four-layer convolutional spiking neural network (CSNN) was trained directly  without any time-consuming preprocessing. This enables efficient adaptation to energy- and resource-constrained on-device SNNs for real-time inference on high-resolution event-based streams.
Compared to recent gesture recognition systems adopting neural networks for vision processing, the proposed neuromorphic vision system achieves competitive accuracy of 94.04\% in a 13-class classification task, and 97.24\% in an unexpected abnormal driver motion classification task with the CSNN architecture. Additionally, when deployed to Intel Loihi 2 neuromorphic chips, the energy-delay product (EDP) of the model achieved 20,721 times more efficient than that of a non-edge GPU, and 541 times more efficient than edge-purpose GPU.
Our proposed CSNN and the dataset can be used to develop safer and more efficient driver-monitoring systems for autonomous vehicles or edge devices requiring an efficient neural network architecture.
\end{abstract}


\section{Introduction}
{W}{ith} the advancement of artificial intelligence (AI), it is being applied across various industrial fields, and among these, vehicle AI systems are emerging as one of the most prominent applications. Onboard AIs in vehicles are used to assist autonomous driving and the safety of drivers and pedestrians by integrating with the control system~\cite{JAS-2019-0401}. In particular, the EU and the United States have introduced mandatory requirements for various driver assistance systems to ensure safe road traffic by regulations on general automotive safety~\cite{eureg}. 
The primary objective of this regulation is to enhance the protection of elderly drivers, vehicle occupants, pedestrians, and cyclists. Given that research shows human error is the cause of 95\% of accidents, implementing this regulation is projected to save more than 25,000 lives and prevent at least 140,000 serious injuries by 2038.
This has led to an urgent need to research and develop driver assistance systems (ADAS) using AI.

Conventional AI vision systems for autonomous vehicles utilize  AI platforms for training with large datasets, after which the learned information is transferred to individual vehicles. These systems are typically implemented using GPU or AI accelerator-based hardware for data training and inference, which result in high power consumption, slow response times, and challenges in real-time prediction~\cite{n1}.
The exponential increase in neural network computations required for massive training datasets in autonomous driving has exacerbated these limitations. To address these challenges, a new generation of low-power, high-performance, and highly efficient AI vision systems is expected to become central to AI implementation in autonomous driving.

Neuromorphic AI vision systems, in particular, represent a way to support AI technology for autonomous vehicles that is gaining increasing significance~\cite{n2}.
Event cameras, also called neuromorphic cameras, and neuromorphic processors, which mimic the functionality of the human eye and brain, offer ultra-low power consumption, high efficiency, and rapid responsiveness. Consequently, these technologies are actively applied in various applications in autonomous vehicles, where they serve as core components for the sensory and processing needs~\cite{n3, massa2020ijcnn, perez2013mapping}.

In response to these demands, we propose N-DriverMotion, an event-based dataset and vision system for neuromorphic learning and predicting driver motions on an efficient convolutional spiking neural network (CSNN). This driver motion recognition research, comprising a neuromorphic framework for efficient CSNN configuration in terms of energy and latency, implements driver safety assistance by incorporating a high-resolution event-based camera. The contributions of the proposed driver motion recognition system include the following:
\begin{itemize}
    \item We create an event-based dataset, N-DriverMotion (720 * 720 resolution), for large-scale driver motion recognition using an event-based camera: To the best of our knowledge, this is the first study for driver motion recognition using a high-resolution event camera and spiking neural networks. The event-based dataset presents 13 driver motion categories classified by direction (front, side), illumination (bright, moderate, dark), and participant.
    \item We present a novel simplified four-layer convolutional spiking neural network (CSNN), directly trained with the event dataset without any time-consuming preprocessing. This enables efficient adaptation to on-device spiking neural networks (SNNs) for real-time inference over event-based camera streams.
    Furthermore, the proposed neuromorphic vision system achieves competitive accuracy of 94.04\% in 13 class classification task, and 97.24$\%$ in unexpected abnormal driver motion classification task with the CSNN architecture, developing safer and more efficient driver monitoring systems for autonomous vehicles or edge devices requiring a low-power and efficient neural network architecture.
    \item We design an efficient CSNN for driver motion recognition for one of the widely used practical neuromorphic frameworks, the Intel Lava neuromorphic framework. We also propose a version of our CSNN for the Loihi 2 processor~\cite{lava, loihi2}. We observed that the energy-delay product (EDP) of the proposed model on Loihi 2 exhibited 20,721 times more efficient than that of a non-edge GPU and 541 times more efficient than that of an edge purpose GPU, which is considered to be one of the most important aspects of edge-device AI.
\end{itemize}

We designed an optimized convolutional spiking neural network (CSNN) for efficient energy use and deployment on on-device applications and systems like Loihi 2. To train the proposed CSNN, we used a surrogate gradient descent-based backpropagation method, SLAYER~\cite{NEURIPS2018_82f2b308}, for direct training with our N-DriverMotion dataset. Additionally, instead of resizing the high-resolution event frames directly, we added a pooling layer to enable adaptive deployment on GPUs and the Loihi 2 system, depending on memory requirements.

The remainder of this paper is organized as follows: Section~\ref{sec:2} briefly reviews related work, while Section~\ref{sec:3} describes the proposed driver motion recognition system. Section~\ref{sec:4} presents the implementation results and performance evaluations and Section~\ref{sec:5} concludes this paper.

\section{RELATED WORK}
\label{sec:2}

Event-based cameras have advantages over frame-based cameras in properties such as high dynamic range, high temporal resolution, low latency, and low power consumption due to their sparse event streams when intensity changes~\cite{9138762}. Therefore, they have been increasingly adopted in machine learning vision systems for gesture recognition and learning.
Hand-gesture recognition~\cite{fnins.2020.00637}, gesture and facial expression recognition~\cite{10298106}, DECOLLE~\cite{10.3389/fnins.2020.00424} for deep continuous local learning, EDDD for event-based drowsiness driving detection~\cite{8990081}, TORE~\cite{9767613} for time-ordered recent events, event-based asynchronous sparse convolutional networks~\cite{10.1007/978-3-030-58598-3_25}, and EST~\cite{9009469} for end-to-end learning of representations for asynchronous event-based data have been developed on deep learning-based neural networks (DNNs) and event representations, exploiting the sparsity and temporal dispersion of event-based gestures. Riccardo et al.~\cite{massa2020ijcnn} trained a deep neural network on a preprocessed DVS hand gesture dataset for recognizing hand gestures on a Loihi neuromorphic processor. The trained DNN was converted into the spike domain for deployment on Intel Loihi~\cite{ref1} for real-time gesture recognition.

Unlike DNN-based approaches, which require converting event-based datasets into static patterns for processing, recent research has shifted toward direct training and learning on SNNs using event-based datasets. This is because events are inherently suited for SNNs operating in continuous time. Moreover, SNNs directly exploit the temporal information of events for energy-efficient computation.
SLAYER proposes an error backpropagation mechanism for offline training of SNNs.
It receives events and leverages the backpropagation method to train synaptic weights and axonal delays directly.
Amir et al.~\cite{amir2017low} proposed an end-to-end event-based gesture recognition system using an event-based camera and a TrueNorth processor configured with a convolutional neural network (CNN). It recognized hand gestures in real time from gesture event streams captured by a Dynamic Vision Sensor (DVS).

Event-based vision systems have been widely used in various applications, including vision systems for autonomous driving~\cite{8578666, 8784777} and object detection~\cite{spikingyolo2020, 10376840}. These applications exploit event-based cameras to capture the spatial and temporal event data of driving and objects.

\begin{figure}[t]
   \centering
   \includegraphics[width=1.1\columnwidth]{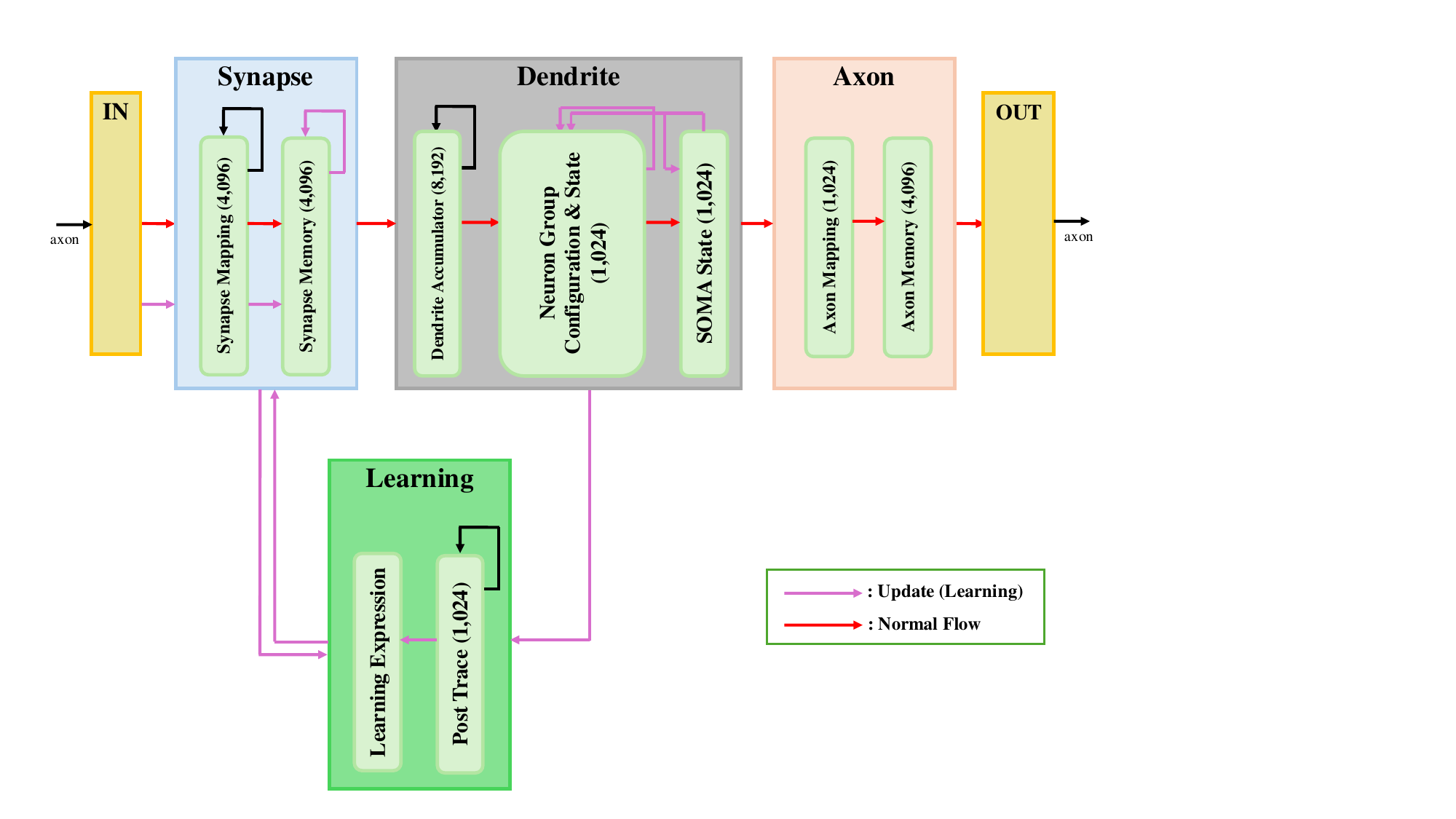}
   \caption{Simplified structure of Loihi which illustrates how neurocore is structured and operates.}
   \label{Loihi_structure}
\end{figure}

Loihi 2 is the second generation of Intel neuromorphic chip designed for energy-efficient AI applications~\cite{loihi2}. Loihi chip applies Current Based Leaky-Integrate and Fire neuron (CUBA LIF), as its main neuron model. The overall behavior of CUBA LIF model is close to that of LIF, but is more biologically plausible as CUBA LIF model includes the temporal dynamics of the postsynaptic input current~\cite{CUBA_LIF}. A single Loihi chip consists of 128 neurocores, which are groups of neurons that function according to the compartments. Each neurocore is able to communicate with other neurocores through the asynchronous Network-on-Chip (NoC) where the messages are packetized for sending and receiving. Although a single chip provides only 128 neurocores, Loihi is capable of transporting information without increasing latency for message exchange between multiple chips. The simplified model of Loihi is shown in Figure~\ref{Loihi_structure}, which illustrates the asynchronous neuron model mechanism, along with the implementation of online learning or a continual learning mechanism. The Loihi chip has been used with the Lava software framework to support the development of agile neuromorphic software and applications~\cite{n4, CarSNN, n5, n6, n7, n8}.

While event-based systems have shown promise in previous research for autonomous driving and gesture recognition, they still face challenges in achieving real-time performance with low power consumption in highly dynamic environments. Additionally, existing systems may struggle to handle the diverse and unpredictable nature of recognition scenarios, particularly when dealing with rapid motions or changing lighting conditions, which can generate massive event streams. Our proposed system with N-DriverMotion addresses these challenges by leveraging the energy-efficient CSNN architecture on Loihi 2. This enables fast, real-time processing of large-scale event data while maintaining low power consumption. Furthermore, the system ensures that event data from event cameras is processed promptly without bottlenecks, which is important for safety-critical applications like autonomous driving.

\section{PROPOSED EVENT-BASED DRIVER MOTION RECOGNITION SYSTEM}
\label{sec:3}

In this section, we introduce efficient CSNN architecture and a driver motion learning and prediction system employing a direct training mechanism and event-based data streams.
As event-based data differs from static images in that it includes time to represent events, we adopt a direct spiking training method and 3D spike convolution operation to build an efficient CSNN model.

For the direct training of spiking neural networks, we exploited a gradient-based training method developed in SLAYER~\cite{NEURIPS2018_82f2b308}. It is generally known that the characteristic of discrete spike events hinders the differentiation in SNN. To resolve such problems, various attempts have been made to propose the approximation for the derivative of the spike function~\cite{7727212, lee2016training, zenke2018superspike}. However, none of the following have considered the temporal dispersion between spikes to resolve the problem. A differentiable approximation (i.e., surrogate gradient) of the spike function is introduced with the probability of a change in the spiking state. It is explained that the derivative of the spike function represents the Probability Density Function (PDF) for the change of state of a spiking neuron.
With an expectation value of the derivative of the spike function, estimated backpropagation error, and  gradient terms for weight and delay in layer $l$, it is possible to get the derivatives of the total loss $E$ as 
\begin{equation}
    \label{eqn2}
    e^{(l)} = 
    \begin{cases}
        \frac{\partial L(t)}{\partial a^{(n_l)}} & \text{if } l = n_l \\
        \left( W^{(l)} \right)^T \delta^{(l+1)}(t) & \text{otherwise}
    \end{cases}
\end{equation}

\begin{equation}
    \label{eqn:label}
    \delta^{(l)}(t) = \rho^{(l)}(t) \cdot (\varepsilon_d \odot e^{(l)})(t)
\end{equation}

\begin{equation}
    \label{eqn3}
    \nabla_{W^{(l)}} E = \int_{0}^{T} \delta^{(l+1)}(t) \cdot \left( a^{(l)}(t) \right)^T \, dt
\end{equation}

\begin{equation}
    \label{eqn4}
    \nabla_{d^{(l)}} E = - \int_{0}^{T} \dot{a}^{(l)}(t) \cdot e^{(l)}(t) \, dt
\end{equation}

where $\rho^{(l)}(t)$ denotes the probability density function in layer $l$ at certain time $t$, $\Delta \xi$ as the random perturbation, $W^{(l)}$ is the weight vector for layer $l$, $a^{(l)}$ being the spike response signal, $L(t)$ being the loss at a certain time $t$, $d^{(l)}$ being the axonal delay, $\varepsilon_d$ being the spike response kernel, and $\odot$ being the element-wise correlation operation in time.
This provides effective distribution of the errors back through layers of a neural network as in DNNs. It takes account of the errors in the previous timeline, a crucial factor to be considered as spiking neuron’s states relied on the previous states.

In our model, the event-based video streams are defined as a 3D tensor with the shape of $(u, v, t)$ where $u$ and $v$ denote the coordinate of the width and height of layer $l$, and $t$ stands as the timestamp~\cite{10.3389/fnins.2020.590164}. By configuring the optimal temporal frequency, users can control $t$ from the original segment.

\begin{figure}[htbp]
   \centering
   \includegraphics[width=1.0\columnwidth]{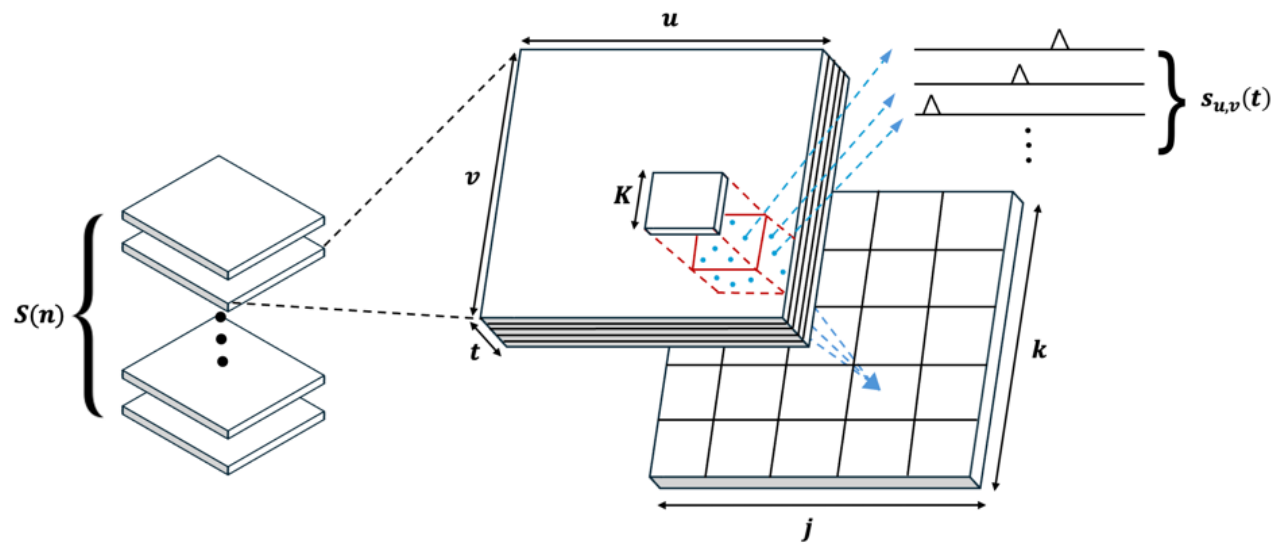}
   \caption{The 3D convolutional spiking operation.}
   \label{convolution}
\end{figure}

A convolutional neural network (CNN) is a feed-forward network composed of multiple layers where filters convolve around the input or single layer for neurons to collect meaningful features or patterns~\cite{forsyth1999object}. However, as the event-based video has not only spatial information but also time, the sampled input sequence $S(n)$ would require a 3D convolutional kernel for building a spiking neuronal feature map as shown in Figure~\ref{convolution}. Each spike inside the kernel represents a cluster of spike trains $s_{u,v}(t)$ where the spike's coordinate is $(u,v)$, and is in range of the temporal resolution window $t$. Spikes from the kernel continually get cumulated to the neurons in the feature map, and when the spikes in the region of the kernel exceed a certain threshold, it would generate the membrane potential for a single neuron in the feature map, creating spatio-temporal dynamic patterns.

The 3D spiking convolution operation is performed by convolving the spike trains in the kernel with a spike response kernel and applying the threshold function. Each spike train is converted into a spike response signal, then subsequently transformed into the membrane potential by integrating the signal with the refractory response of the neuron:

\begin{equation}
    \label{eqn5}
    a_{u,v}(t) = S_{u,v}(t) * \varepsilon_d(t)
\end{equation}

\begin{equation}
    \label{eqn6}
    u_{j,k}(t) = \sum_{m=1}^{K} \sum_{n=1}^{K} W_{m,n} a_{m+(j-1),n+(k-1)}(t) + (S_{j,k}(t) * \nu(t))
\end{equation}

\begin{equation}
    \label{eqn7}
    S_{j,k}(t) = 1 \ \text{and} \ u_{j,k}(t) = 0 \quad \text{when} \quad u_{j,k}(t) \geq V_{\text{thr}}
\end{equation}

where $*$ denotes the convolution operator, $W_{m,n}$ denotes the synaptic weights at $(m,n)$ of the kernel, $u_{j,k}$ represent the membrane potential at coordinate $(j,k)$ presented by the feature map, $K$ represents the width and height of the convolution kernel, and $\nu(t)$ is the refractory kernel.

\begin{figure*}[htbp]
   \centering
   \includegraphics[width=1.0\columnwidth]{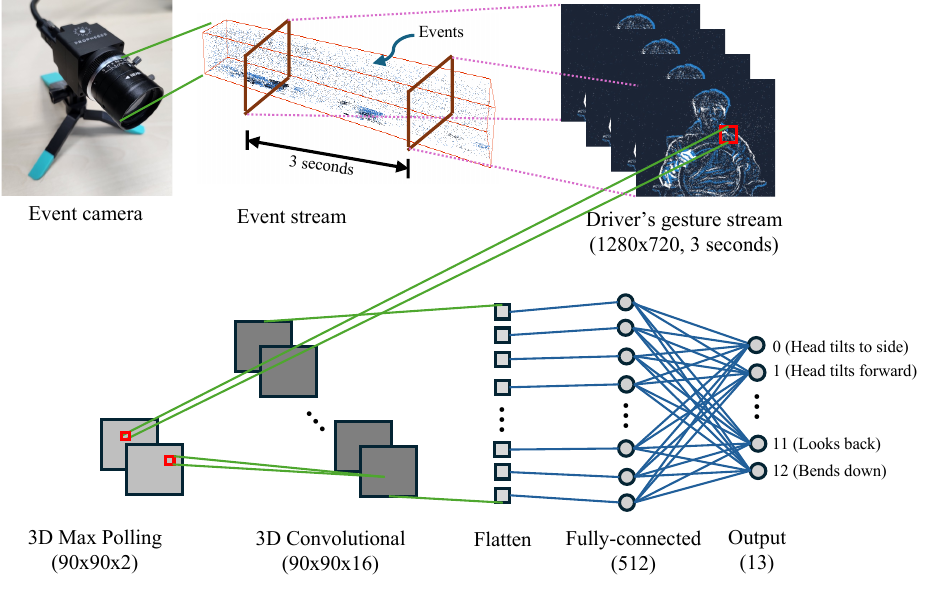}
   \caption{Proposed Event-based Driver Gesture Recognition System.}
   \label{system}
\end{figure*}

Our proposed model is developed based on existing models~\cite{amir2017low, NEURIPS2018_82f2b308} for DVS Gesture recognition, which implements SNN training and a 3D spike convolution operation. The SLAYER and TrueNorth models consist of 8 layers and 16 layers, respectively. Figure~\ref{system} presents our proposed network model and system, where the network is simplified and comprises only 4 layers. When tested with a more complex model, we observed that it not only occupied more memory, but also had a detrimental impact on overall performance, as more layers hindered the neurons from firing spikes, leading to lower event rate because spikes could not be passed to consecutive neuron layers.

The 720x720 resolution event streams for driver motion recognition were recorded by an event-based camera for 3 seconds and delivered to the first 3D pooling layer. Unlike conventional convolutional neural networks, where convolution layers are processed first, our model implements a pooling layer before passing the data to the convolution layer. The main reason for this is to down-sample feature maps to resolve large memory usage, as well as to extract features. Compared to classical pooling layers, SNN pooling layers are considered  trainable layer, as they not only down-sample the event streams but also take into account temporal aspects, enabling the learning of spatio-temporal features from the input. The pooling layer uses a kernel of size 8x8 and passes data to the convolution layer, which uses a 5x5 kernel to produce 16 channels of features. The features are then fed to 2 final fully-connected layers, which provide 13 outputs representing the spike rates for the given event data.

We notice that the existing spiking neural networks designed for other applications are not optimized for autonomous driving, which requires low power consumption, real-time inference, and efficient deployment on actual neuromorphic chips. To effectively implement driver motion recognition for autonomous driving, it is important to have a neural network that is optimized for neuromorphic hardware (simple but without performance degradation) as well as an event-based dataset for training such networks. We addressed these elements by designing, implementing, and testing a network that takes these features into consideration.

\section{EXPERIMENTS AND RESULTS}
\label{sec:4}

\subsection{EVENT-BASED CAMERA AND DATA CONVERSION}

Prophesee's Metavision EVK4\footnote{EVK4 camera is an HD Event-Based Vision evaluation kit mounted with IMX636 sensor, developed by Sony and Prophesee. Please refer to https://www.prophesee.ai/event-camera-evk4/} camera is one of the latest event-based vision sensors that supports up to 1280x720 pixel resolution. Generally, when a change in pixel values exceeding a certain threshold defined by the user occurs, the event-based camera asynchronously detects such changes in brightness and generates events specific to that pixel~\cite{9138762}. Each event contains pixel information describing the position in the x and y coordinates, indicating a motion's change, and a timestamp for the occurrence of the event. The device offers a high dynamic range (86 dB, however, it can reach over 120 dB based on low light cutoff measurement being: 0.08 lux) and a typical and maximum event rate of 1.06 giga-events per second (Geps).

In the experiment, in order to allow direct usage of the spike data obtained from the high-resolution event-based camera without requiring additional resources, we minimized modification to the extractor so that it can retrieve coordinate, polarity, origin (a newly implemented feature in the EVK4), and timestamp information. Moreover, we observed that the most meaningful motion information was contained in the central 720x720 pixels so that the size of the input events was cropped to 720x720 pixels.

\subsection{DATASET}
Over the past years, a profusion of gesture datasets captured with simple frame-based sensors has been presented. However, to stimulate the improvement of event-based computer vision, Hu et al.~\cite{hu2016dvs} strongly support the importance of the DVS dataset. Serrano et al.~\cite{serrano2015poker} introduced the first labeled event-based neuromorphic vision sensor dataset which originated from the classical MNIST digit recognition dataset by moving the images along certain directions within the screen. This dataset was further developed by Orchard et al.~\cite{orchard2015converting}, who removed several frame artifacts using a pan-tilt unit. It can be observed that generating artificial movement for static images enabled the production of event outputs, which became important for object recognition in neuromorphic systems such as spiking neural networks~\cite{perez2013mapping}. However, it has been observed that static image recognition with event-based vision sensors is ineffective, as their main purpose is focused on dynamic scenes. Recognizing these drawbacks, some datasets consisting of dynamic scenes have been presented. Berner et al.~\cite{berner2013240} introduced novel datasets that converted the existing visual video benchmarks for object tracking and action/object recognition into spiking neuromorphic datasets using the DAVIS camera's output.

\begin{figure*}[htbp]
   \centering
   \includegraphics[width=0.9\columnwidth]{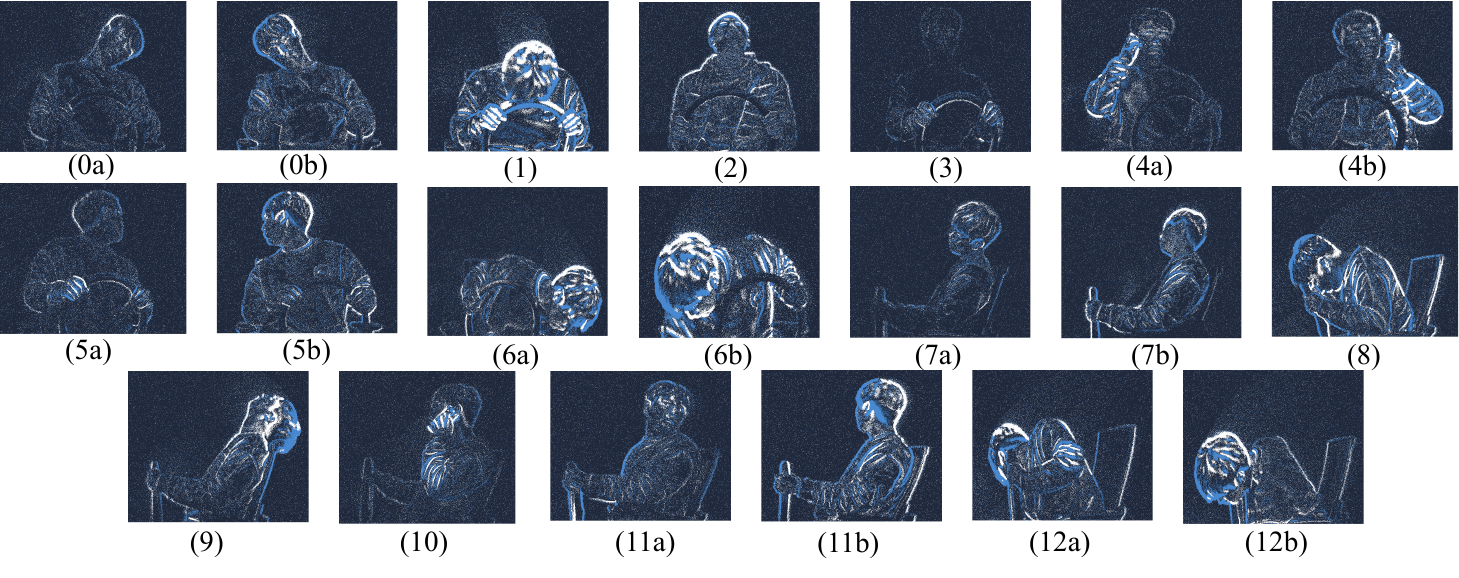}
   \caption{Sample shots taken for each driver gesture.}
   \label{gestures}
\end{figure*}

\begin{table}[htbp]
\caption{Labels of the corresponding driver gestures} 
\begin{center}
\begin{tabular}{|c|c|}
\hline
Label & Gesture\\\hline
\multicolumn{2}{|c|}{Front} \\\hline
0 & Head tilts to side (a: left, b: right) \\
\hline
1 & Head tilts forward \\
\hline
2 & Head tilts backward \\
\hline
3 & Normal driving \\
\hline
4 & Phone call (a: right, b: left) \\
\hline
5 & Looks back (a: left, b: right) \\
\hline
6 & Bends down (a: left, b: right) \\\hline
\multicolumn{2}{|c|}{Side} \\\hline
7 & Head tilts (a: left, b: right) \\
\hline
8 & Head tilts forward \\
\hline
9 & Head tilts backward \\
\hline
10 & Phone call \\
\hline
11 & Looks back (a: left, b: right) \\
\hline
12 & Bends down (a: left, b: right) \\
\hline
\end{tabular}
\end{center}
\label{table_gestures}
\end{table}

Even though these datasets show dynamic movements in a spiking neuromorphic way, according to Tan et al.~\cite{tan2015benchmarking}, a DVS camera produces microsecond temporal resolution output. In contrast, datasets produced by conventional vision tools such as video recorders or color cameras have tens of milliseconds of temporal resolutions, which causes a loss of high temporal frequency during conversion. Moreover, they add that there is a high chance of subsidiary unwanted artifacts being present during conversion. To address these problems, Amir et al.~\cite{amir2017low} present the DvsGesture dataset, which directly captured scenes with the DVS128 camera. Additionally, the DvsGesture datasets were captured under different lighting conditions, as DVS sensors are less affected by brightness and introduce some meaningful noise. However, the DvsGesture dataset is mainly utilized for simple classification tasks, which may be considered less practical.

The N-DriverMotion dataset comprises 1,239 instances of a set of 13 driver motions (1 normal driving motion and 12 driving motions that are considered possibly dangerous) as shown in Figure~\ref{gestures} and Table~\ref{table_gestures}. These motions include head falling (front, sides, and back), holding a cellphone (right/left hand), looking back (to the right/left side), the body going down (to the right/left side), and all the same motions above in side-captured manner. The dataset was collected from 23 subjects under three different lighting conditions. All subjects were within a specific age range; they had no disabilities, or any other problems that could affect driving. Each subject performed all 13 motions in a single recording, where each motion was recorded for around 3 seconds under the same brightness condition. The three brightness conditions were classified as "bright," "moderate," and "dark" to test how much brightness would affect the recordings and to examine the robustness of event data in harsh illumination condition. We note that informed consent was obtained from all participants during data collection.
To evaluate the classifier's performance, we randomly selected 992 motions for the training set, and the remaining 247 motions were designated as the test set.

Due to limitations in recruiting participants, we were unable to include experiments involving drivers with medical conditions or disabilities. This will be expanded in future work. Instead, to address unpredictable scenarios, we introduced five new types of event data that were not used during training. These additional test cases include abnormal reactions such as seizures during driving, sudden standing or rising movements, collisions where the driver hits the windshield, hand-swipe gestures, and abnormal camera angles caused by accidents or issues with the camera's fixed position. These newly created event-based video sequences were specifically designed to test how the system responds to other abnormal cases not seen during training.

\subsection{EXPERIMENTAL SETUP AND TOOL FLOW}

The experiment is mainly consisted of 3 sub-parts to measure different performance of our proposed model: the accuracy of the 13-class driver motion classification, the accuracy of classifying unexpected abnormal driver motion versus normal driver motion, and the energy efficiency with respect to throughput in various AI accelerators.

\subsubsection{Multiple driver motion classification}
The aim of this experiment lies in observing how accurately our proposed system can classify various driver motions (13 classes in total). With the model developed based on the Lava-DL API, our proposed network will be loaded onto the RTX 3080 GPU and trained for 200 epochs. During each epoch in training, the models would be evaluated with the test set and produce accuracy based on the spike rates of the output spikes. If the model produces the highest test accuracy, the model's weights are stored as a PyTorch state dictionary file. Further explanation about the configuration of the system is shown in subsection~\ref{subsec:training}.

\begin{figure}[t]
   \centering   \includegraphics[width=0.8\columnwidth]{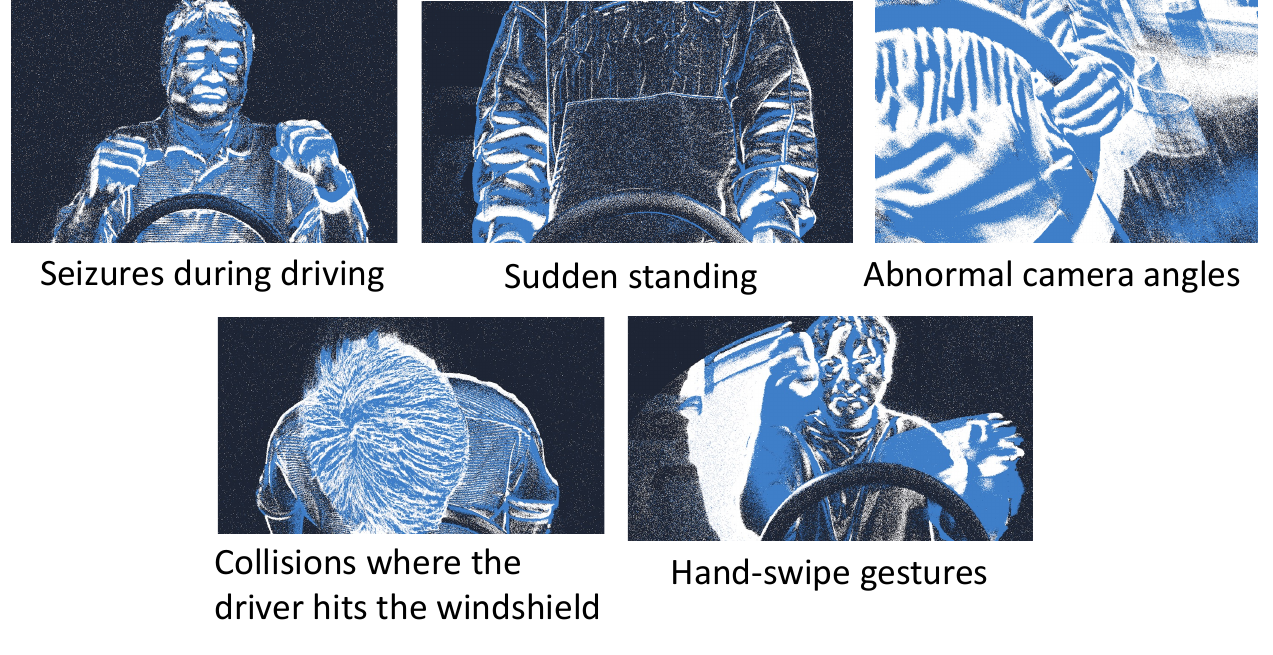}
   \caption{Sample shots taken of abnormal driver gestures. Note that hand-swipe gestures are not simple 
   hand-waving actions but are actions to protect oneself from various attacks}
   \label{abnormal}
\end{figure}

\begin{table}[t]
\caption{Abnormal driver gestures} 
\begin{center}
\begin{tabular}{|c|}
\hline
Seizures during driving \\
\hline
Sudden standing  \\
\hline
Collisions where the driver hits the windshield \\
\hline
Hand-swipe gestures \\
\hline
Abnormal camera angles \\
\hline
\end{tabular}
\end{center}
\label{table_gestures2}
\end{table}

\subsubsection{Inference on unseen abnormal driving motions}
While it is important to see how the model performs classifications based on the provided 13 motion sets, it is also important to check the model's capability in classifying unexpected motion sets. Here, we make inferences on a separate dataset that is composed of abnormal driver motions not included in the 13 classes, as well as normal motions, where the model has been trained to classify them as label 3. In this experiment, we have set the objective as binary classification: abnormal driver motion and normal driving motion. Given such abnormal driver motions, if the model produces predictions other than label 3 (normal driving motion), we can observe that the model is capable of classifying them as abnormal motion, and vice versa. We have used the fully trained CSNN model with the same configuration for event data (e.g., sampling time) that was used for measuring accuracy on the 13-class classification task. The test data for this task is composed of 82 abnormal driver motions (10 participants performing 5 different unexpected abnormal driving motions, as shown in figure~\ref{abnormal} and table~\ref{table_gestures2}) along with 63 normal driver motions from the original test set. The evaluation process was conducted in the same way as the multiple driver motion classification, except in a binary classification fashion.

\subsubsection{Energy efficiency in various AI accelerators}
Energy efficiency is one of the most promising aspects of SNN, and this experiment is designed to show the overall efficiency of the model with respect to 3 AI accelerators: Nvidia RTX 3080, Nvidia Jetson Xavier NX, and Intel Loihi 2. For both Nvidia GPUs, we loaded the models directly onto the devices and measured various inference costs (latency, total energy consumption, etc.). Due to fast inference speed of GPUs, we measured by averaging each of the total costs by the number of samples processed to measure the values. For Loihi 2, we designed a specific pipeline that enables us to load the model into the system and feed the event data directly to the model, along with the utilization of measurement tools provided by the Lava API. Detailed information about experimental flows and settings for Jetson Xavier NX and Loihi 2 is described in subsection~\ref{subsec:edge}.

\subsection{NETWORK CONFIGURATION AND TRAINING}
\label{subsec:training}

To enable adequate training, the dataset was converted from binary format into tensors consisting of x-coordinates, y-coordinates, polarities, and timestamps. We also set the sampling time to 2.0 seconds, as the core movements were mostly within this range. More importantly, we wanted to check whether our proposed model is capable of classifying driver motions in a short amount of time, as in real-time applications, such systems should detect abnormal activities rapidly. In this form, the data can now be used as input to the CSNN during runtime.

All of the applications were implemented and mapped to GPU/CPU to be executed as the simulation for Loihi 2, using the Lava version 0.4.4 software framework and the Lava-DL library for training.
Our CSNN models were developed with neuron models and trained with the event-based dataset using the SLAYER API, which were all provided by the Lava-DL.

We trained multiple CNN configurations to compare accuracy results. Each network was trained for 200 epochs with a single batch size due to the dataset's high resolution. We selected CUBA LIF as the base neuron model for the spiking CNN models. For optimization, the learning rate was set to 3$\times$10\textsuperscript{-3} and ADAM was used as the optimizer. For the loss function, we chose spike rate loss. The network configuration for the 4-layered CSNN and the neuron parameters are shown in Table~\ref{table_network}

\begin{table}[htbp]
    \scriptsize
    \begin{minipage}[t]{0.5\textwidth}
        \centering
        \begin{tabular}{l|lllll}
            & layer   & map size  & features & kernel & padding \\ \hline
            & N/A     & 720 x 720 & 2        & N/A    & N/A     \\
            1 & Pool    & 90 x 90   & 2        & 8     & N/A     \\
            2 & Conv    & 90 x 90   & 16       & 5      & 2       \\
            3 & Flatten & 90 x 90   & 16       & N/A    & N/A     \\
            4 & Dense   & 512       & N/A      & N/A    & N/A     \\
            5 & Dense   & 13        & N/A      & N/A    & N/A    
        \end{tabular}           
    \end{minipage}%
    \begin{minipage}[t]{0.5\textwidth}
        \centering
        \begin{tabular}{l|l}
            Neuron Parameters & Value \\
            \hline
            Voltage threshold & 1.25 \\
            Current decay & 0.25 \\
            Voltage decay & 0.03 \\
            Tau gradient & 0.03 \\
            Scale gradient & 3 \\
            True rate & 0.2 \\
            False rate & 0.03 \\
        \end{tabular}
    \end{minipage}
    \caption{Network configuration of 4-layered CSNN(left) with neuron parameters(right)}
    \label{table_network}
\end{table}

\subsection{EDGE DEVICE CONFIGURATIONS AND INFERENCE}
\label{subsec:edge}
In this section, we provide a detailed description of various AI accelerators that are specialized for use in edge devices. We utilized two different edge AI accelerators: Jetson Xaior NX, and Intel Loihi 2 to compare overall energy efficiency with respect to image processing delays.

\subsubsection{Jetson Xavior NX}
The Jetson Xavior NX is a fully featured development board equipped with a 6-core ARM CPU and a 48 tensor core processing unit supporting a clock frequency up to 1.1 GHz, along with 16 GB of memory. To minimize power usage, we operated on the standard module, where the GPU’s clock frequency is set to 0.67 GHz. All inference configurations were the same as with the RTX 3080, except for the library version. Due to the restriction of the Python version, inference on the Jetson Xavior NX was executed with a lower version of the lava-dl (0.4.0) library.

\begin{figure*}[t]
   \centering
   \includegraphics[width=1.0\columnwidth]{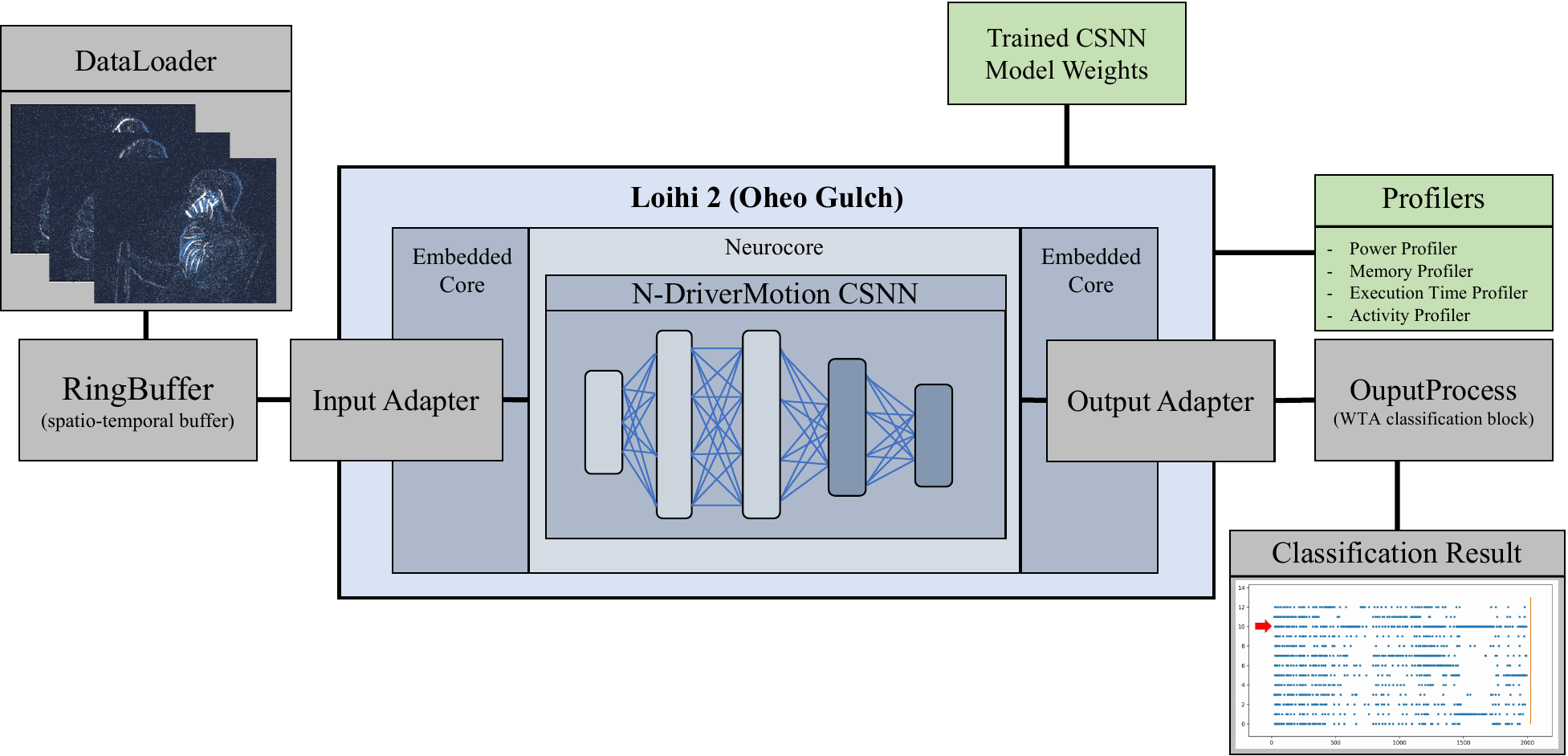}
   \caption{Pipeline for Inference of CSNN on the Loihi 2 processor. The blocks that are part of the main processes are colored as gray, while the green blocks are not part of the main process.}
   \label{loihi_pipeline}
\end{figure*}

\begin{table}[t]
\caption{Network configuration of a 4-layered CSNN for Loihi 2}
\begin{center}
\begin{tabular}{|l|l|l|l|l|}\hline
layer   & map size  & channels & kernel & padding \\ \hline
Input    & 40 x 40  & 2        & N/A    & N/A     \\ \hline
Conv    & 40 x 40   & 4       & 5      & 2       \\ \hline
Flatten & 40 x 40   & 4       & N/A    & N/A     \\ \hline
Dense   & 512       & N/A      & N/A    & N/A     \\ \hline
Dense   & 13        & N/A      & N/A    & N/A    \\ \hline
\end{tabular}
\end{center}
\label{table_network3}
\end{table}

\subsubsection{Intel Loihi 2}
A Neuromorphic chip is one of the edge AI accelerators that particularly targets the implementation of SNNs. Among various neuromorphic chips, we used Loihi 2 to observe the overall power consumption of the benchmark three fully connected layer SNN model, and the proposed CSNN model, as Loihi 2 adopts CUBA LIF as the base neuron model best fitting all of the proposed models~\cite{CarSNN}. For this experiment, we used the Oheo Gulch, one of the Loihi 2-based neuromorphic systems that contains a single-socketed Loihi 2 chip. Due to the limited availability of Loihi 2 chips, along with the nature of large input and model size, some adjustments to the input and models were made. We downsampled the input from 720x720 to 40x40 with max pooling and extracted the motion stream time from total of 3.0 seconds to 1.7 seconds. For the three fully connected layer model, there were no changes. For the proposed CSNN model, we excluded the CUBA pooling layer since the input had already been downsampled and reduced the number of features extracted from 16 to 4. The description of the network model on Loihi 2 is shown in Table~\ref{table_network3}. We were able to load the three fully connected layer model onto a single Loihi 2 chip, whereas the proposed CSNN model required a total of 4 Loihi 2 chips to be loaded.

To enable inference on Loihi 2, it is necessary to create a pipeline that can process spatio-temporal data for classification tasks on Loihi 2. We created a pipeline to conduct inference on Loihi 2, which is shown in Figure~\ref{loihi_pipeline}. The event data was stored in a buffer block named RingBuffer, where it passed the frame data one at a time to Loihi 2. As the frame data was passed to Loihi 2 the last layer in the model produced spikes that generated the classification result for the single frame. The spikes were cumulatively stored in the OutputProcess block until every frame in the event data was processed. The final classification was conducted in a Winner-Takes-All (WTA) fashion, where the class with the most spikes was the predicted value. 

The total description of how the models were converted and converged at the Oheo Gulch neuromorphic hardware is shown in Table~\ref{3FCN} and Table~\ref{CSNN_partition_core_usage}, where each column indicates the percentage of the total usage for the input axons, neuron group, neurons, synapses, axonal mapping, and axon memory, respectively.

\begin{table*}[htbp]
\caption{Resource utilization for a 3-fully connected layer model on Loihi 2}
\centering
\begin{tabular}{|c|c|c|c|c|c|c|c|}
\hline
\textbf{AxonIn} & \textbf{NeuronGr} & \textbf{Neurons} & \textbf{Synapses} & \textbf{AxonMap} & \textbf{AxonMem} & \textbf{Total} & \textbf{Cores} \\ 
\hline
3.20\% & 12.50\% & 0.32\% & 12.80\% & 0.08\% & 0.00\% & 12.94\% & 1 \\ 
3.20\% & 12.50\% & 1.56\% & 73.60\% & 0.40\% & 0.00\% & 62.09\% & 8 \\ 
25.31\% & 12.50\% & 0.12\% & 50.62\% & 0.03\% & 0.00\% & 60.80\% & 103 \\ 
\hline
\textbf{Total} &  &  &  &  &  &  & \textbf{112} \\
\hline
\end{tabular}
\label{3FCN}
\end{table*}

\begin{table*}[htbp]
\caption{Resource utilization for the proposed CSNN model on Loihi 2. The amount of core utilization for the proposed model is more than twice of the 3-fully connected layer model due to the convolutional layer acting as the bottleneck of the model}
\centering
\begin{tabular}{|c|c|c|c|c|c|c|c|}
\hline
\textbf{AxonIn} & \textbf{NeuronGr} & \textbf{Neurons} & \textbf{Synapses} & \textbf{AxonMap} & \textbf{AxonMem} & \textbf{Total} & \textbf{Cores} \\ 
\hline
1.60\% & 12.50\% & 0.32\% & 6.40\% & 0.08\% & 0.00\% & 6.54\% & 1 \\ 
40.00\% & 12.50\% & 0.02\% & 40.00\% & 0.01\% & 0.00\% & 64.02\% & 256 \\ 
0.01\% & 12.50\% & 62.52\% & 0.11\% & 16.00\% & 0.00\% & 25.70\% & 3 \\ 
\hline
\textbf{Total} &  &  &  &  &  &  & \textbf{260} \\
\hline
\end{tabular}
\label{CSNN_partition_core_usage}
\end{table*}

\subsection{Results}

\begin{figure*}[htbp]
   \centering
   \includegraphics[width=1.0\columnwidth]{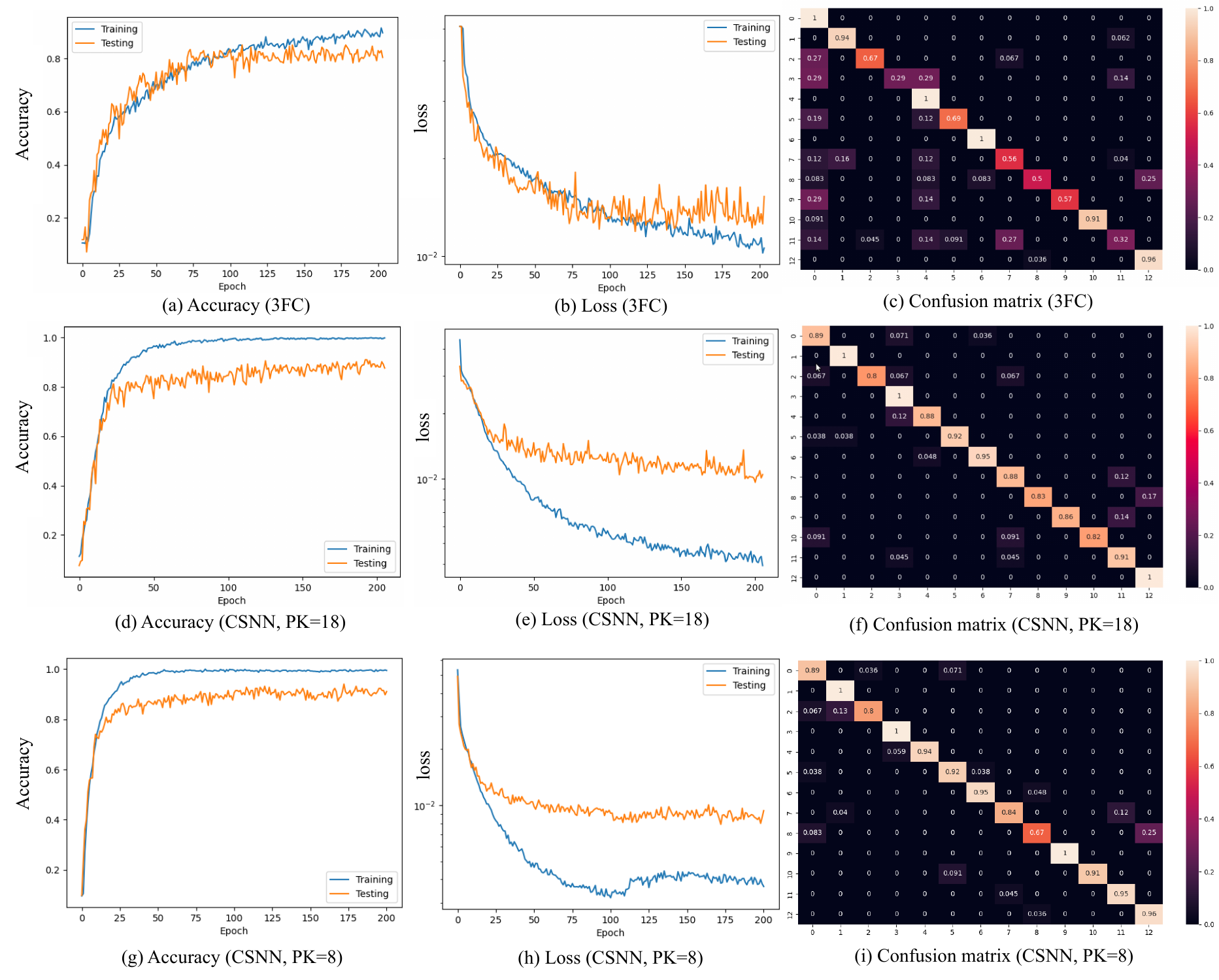}
   \caption{Results (accuracy, loss, and confusion matrix) of the proposed CSNN with pooling kernel sizes 8 and 18 (PK=8 and PK=18) vs. three fully-connected layer SNN (3FC).}
   \label{result}
\end{figure*}

\begin{table}[b]
\centering
\scriptsize
\resizebox{\columnwidth}{!}{%
\begin{tabular}{|c|c|c|c|}
\hline
\textbf{Dataset} & \textbf{Method} & \textbf{Architecture} & \textbf{Accuracy} \\
\hline\hline
\multirow{2}{*} & TrueNorth & SNN (16 layers) & 91.77\% (94.59\%)\\
\cline{2-4}
{DVS Gesture} & SLAYER & SNN (8 layers) & \textbf{93.64 ± 0.49\%} \\
\cline{2-4}
& SLAYER & Proposed CSNN (4 layers) & 92.80\% \\
\hline
\multirow{2}{*} & SLAYER & SNN (3 fully-connected layers) & 85.11\% \\
\cline{2-4}
{N-DriverMotion} & SLAYER & Proposed CSNN (4 layers, 
 pooling kernel size: 8) & \textbf{94.04\%} \\
\cline{2-4}
& SLAYER & Proposed CSNN (4 layers, pooling kernel size: 18) & {91.07\%} \\
\hline
\end{tabular} 
}
\caption{Classification results comparing DVS gesture dataset and N-DriverMotion dataset. For DVS Gesture, it includes results from both SLAYER and IBM TrueNorth.}
\label{accuracy_result}
\end{table}

\begin{figure*}[htbp]
   \centering
   \includegraphics[width=1.0\columnwidth]{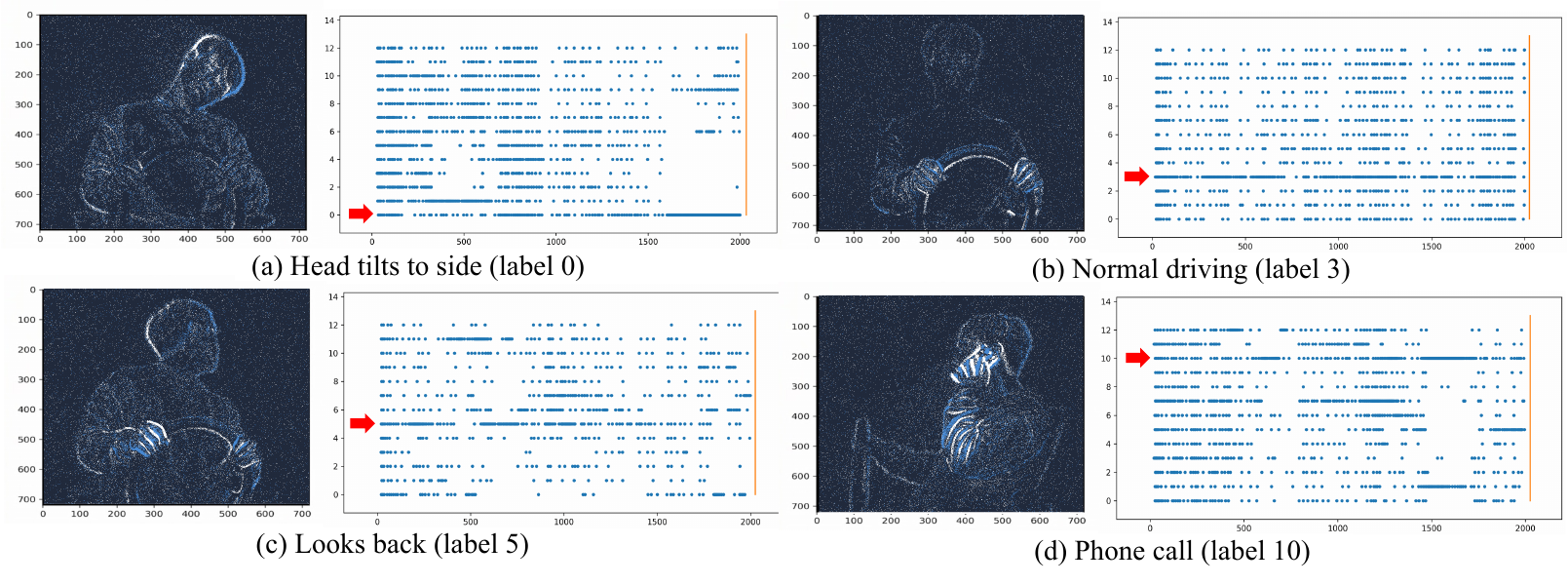}
   \caption{Classification result with output spikes}
   \label{output_spikes}
\end{figure*}

\subsubsection{Accuracy on 13-class prediction}
The results for 13 motion classifications from the training models are shown in Figure~\ref{result} and Table~\ref{accuracy_result}. In the benchmark dataset for evaluating our models, we used the DVS Gesture recognition dataset, comprising 11 hand gesture categories from 29 subjects under 3 illumination conditions~\cite{amir2017low}. We tested three different SNN configurations: 1) all layers composed of dense connection, 2) a convolutional spiking layer with 2 fully connected layers, with the pooling kernel size set to 8, and 3) the same model as 2, except with the pooling kernel size set to 18. For all models, we selectively used only the first 2.0 seconds out of 3.0 seconds of motion streams for each class to classify the actions.
Additionally, the temporal resolution was set to 5 milliseconds to increase training efficiency. For each experiment, we recorded the best loss and accuracy of the SNNs for both training and inference steps, as well as the confusion matrix for the corresponding inference results. The confusion matrix in Figure~\ref{result} shows that the normal situation (label 3) is distinguished from the hazardous situations.
Figure~\ref{output_spikes} illustrates the classification results with output spikes captured on the output layer, classifying 13 driver motions.

The system from the proposed CSNN model with 4 layers showed the highest accuracy, reaching 94.04\% in the test. The inclusion of the pooling and convolution layers improved accuracy performance by almost 8\% over the 3-layered model, which was composed only of fully connected layers. Even though our model is simplified with restrictions applied to resolve the out-of-memory problem caused by the high-resolution event data for training and testing, it was able to demonstrate comparable results. We note that our proposed dataset has more categories for classification and does not include repetitive actions, unlike the DVS Gesture recognition dataset.

\subsubsection{Accuracy on unexpected abnormal driver motion classification task}
\begin{figure}[t]
   \centering
   \includegraphics[width=0.7\columnwidth]{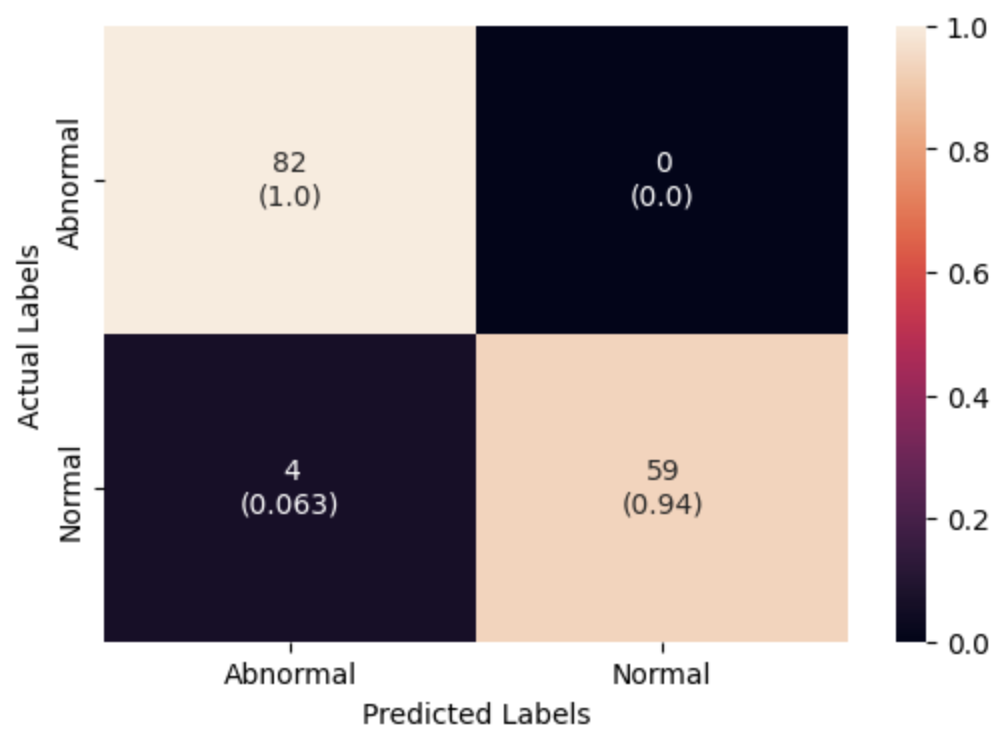}
   \caption{Classification result for unexpected abnormal driver motion. The model detects all the abnormal actions correctly, demonstrating high sensitivity in detecting these actions, and also classifies normal driver motions with high accuracy.}
   \label{confusion_matrix_abnormal}
\end{figure}

The results for binary classification of unexpected abnormal motions and normal motions are displayed in Figure~\ref{confusion_matrix_abnormal}. Except for 4 normal motions which were classified as abnormal driving motions, the model was highly capable of distinguishing between two categories, reaching an accuracy of 97.24\%. One significant point to note is that there were no abnormal motions misclassified as normal motions, which demonstrates the robustness of our proposed CSNN model in detecting such unexpected motions. 

\subsubsection{Latency and Energy}
To measure the power consumption for the Nvidia RTX 3080, we used the NVIDIA Management Library (NVML), which allows us to extract the throughput and energy consumption with respect to single event inference. Due to the fast inference time of the GPU, we calculated the latency by averaging the values, dividing the overall inference time by the total number of inference samples. Energy consumption was also measured by averaging the total energy over the total time of inference.

Nvidia Jetson Xavier NX does not support NVIDIA Management Library, unlike the RTX 3080 GPU. To resolve this issue, we utilized jetson$\_$stats library, which enables us to keep track of the energy consumptions of the GPU by logging the statistics as a CSV file. Similar to the RTX 3080, the calculation for latency was done by averaging the values. The experiment was executed with no additional GPU processes other than the measurement process being loaded.

In the case of Loihi 2, the LAVA framework supports profiler tools which enables us to obtain various measurement values related to power, execution time, counts of neurocore activities, and neurocore SRAM utilization rates. To get precise latency and energy dissipation results, we ran the model 1 million times, which provided enough time for the profiler to measure the overall costs. All experiments for latency and energy comparisons between the GPUs and Loihi 2 were conducted with equal models, both adjusted for Loihi 2. The results for throughput and various energy-related measurements for single sample inference are shown in Table~\ref{energy result}.

The overall energy consumption for Loihi 2 for 1 million inferences was 1.82 J which was highly comparable to that of both GPUs, where the RTX 3080 required over 500 J, and the Jetson Xavior NX required 1.13 J for a single inference on the CSNN model. The throughput, however, showed that while Loihi 2 processed 40.835 samples per seconds, the RTX 3080 processed 66.214 samples per seconds. We believe that this was due to the inference method of Loihi 2, which did not use matrix multiplication, but only addition for the input spike data. The Jetson Xavior NX performed very poorly on throughput, with only 5.46 samples being processed per second.

To compare the balance of low energy and fast runtimes of the devices, we calculated the energy-delay product (EDP), where the energy consumption and performance of the devices were weighted equally~\cite{EDP}. The calculation for EDP was conducted by multiplying the total energy by the latency for a single sample inference. In the EDP comparison between Loihi 2 and the two GPUs, we observed that Loihi 2 had a better EDP than the GPUs for both models, with the 3FCN model being 1.8 million times and the CSNN model being 20,721 times more efficient compared to the RTX 3080, and the 3FCN model being 16,278 times and the CSNN model being 541 times more efficient than the Jetson Xavior NX. 

\begin{table}[t]
\centering
\caption{Comparisons between GPUs and Loihi 2 on inference cost for a single sample. It is clear that despite the throughput of the RTX 3080 being lower, the energy delay product (EDP) of Loihi 2 is much lower than that of both GPUs, demonstrating the high energy efficiency of Loihi 2 for SNN models}
\scalebox{0.9}{
\begin{tabular}{|l|c|ccccc|}
\hline
\multirow{3}{*}{} &
  \multirow{3}{*}{Hardware} &
  \multicolumn{5}{c|}{Inference Cost Per Single Sample} \\ \cline{3-7} 
 &
   &
  \multicolumn{1}{c|}{Latency} &
  \multicolumn{1}{c|}{Throughput} &
  \multicolumn{2}{c|}{Energy} &
  EDP \\ \cline{3-7} 
 &
   &
  \multicolumn{1}{c|}{(ms)} &
  \multicolumn{1}{c|}{(sample/s)} &
  \multicolumn{1}{c|}{Total (mJ)} &
  \multicolumn{1}{c|}{Dynamic(mJ)} &
  $\mu$Js \\ \hline
\multicolumn{1}{|c|}{\multirow{3}{*}{NDriverMotion 3FCN}} &
  Nvidia RTX 3080 &
  \multicolumn{1}{c|}{11.544} &
  \multicolumn{1}{c|}{86.627} &
  \multicolumn{1}{c|}{5.163 x $10^5$} &
  \multicolumn{1}{c|}{N/A} &
  5.94 x $10^6$ \\ \cline{2-7} 
\multicolumn{1}{|c|}{} &
  \multicolumn{1}{l|}{Nvidia Jetson Xavier NX} &
  \multicolumn{1}{c|}{83.3} &
  \multicolumn{1}{c|}{12.011} &
  \multicolumn{1}{c|}{6.29 x $10^2$} &
  \multicolumn{1}{c|}{N/A} &
  5.24 x $10^4$ \\ \cline{2-7} 
\multicolumn{1}{|c|}{} &
  Loihi 2 Oheo Gulch &
  \multicolumn{1}{c|}{1.33 x $10^3$} &
  \multicolumn{1}{c|}{442.364} &
  \multicolumn{1}{c|}{2.42 x $10^{-3}$} &
  \multicolumn{1}{c|}{1.89 x $10^{-4}$} &
  \textbf{3.219} \\ \hline
\multirow{3}{*}{NDriverMotion CSNN} &
  Nvidia RTX 3080 &
  \multicolumn{1}{c|}{15.103} &
  \multicolumn{1}{c|}{66.214} &
  \multicolumn{1}{c|}{5.218 x $10^5$} &
  \multicolumn{1}{c|}{N/A} &
  7.88 x $10^6$ \\ \cline{2-7} 
 &
  \multicolumn{1}{l|}{Nvidia Jetson Xavier NX} &
  \multicolumn{1}{c|}{1.83 x $10^2$} &
  \multicolumn{1}{c|}{5.4617} &
  \multicolumn{1}{c|}{1.13 x $10^3$} &
  \multicolumn{1}{c|}{N/A} &
  2.06 x $10^5$ \\ \cline{2-7} 
 &
  Loihi 2 Oheo Gulch &
  \multicolumn{1}{c|}{1.441 x $10^4$} &
  \multicolumn{1}{c|}{40.835} &
  \multicolumn{1}{c|}{2.64 x $10^{-2}$} &
  \multicolumn{1}{c|}{7.46 x $10^{-4}$} &
  \textbf{380.29} \\ \hline
\end{tabular}%
}
\label{energy result}
\end{table}

\subsection{FUTURE WORK}
\label{sec:5}
While our proposed driver motion recognition system achieves significant improvements in energy efficiency and accuracy, there are several directions for future work to further enhance its performance and applicability. First, we plan to extend the N-DriverMotion dataset by increasing the diversity of driving scenarios, including more complex driving conditions, varied motion patterns, and a larger pool of participants. This will enable the model to generalize more effectively to real-world driving environments. Additionally, we plan to explore approaches for enhancing safety for pedestrians and vehicles in autonomous driving conditions. Another important area of exploration is the integration of online learning mechanisms within the neuromorphic system, allowing the model to adapt dynamically to changing driving conditions and new motion patterns in real time. To achieve this, we will explore unsupervised learning methods based on the spike-timing-dependent plasticity (STDP) learning rule. Moreover, we will evaluate the system's scalability and performance across different hardware platforms, including more neuromorphic processors and edge devices. Lastly, we aim to optimize the neuromorphic pipeline to address challenges such as latency reduction and increased robustness in highly dynamic and unpredictable driving scenarios, further enhancing the system's practicality in real-time autonomous driving applications.

\section{CONCLUSION}
\label{sec:6}
In this paper, we propose N-DriverMotion, a newly collected event-based dataset designed to learn and predict driver motions using a neuromorphic vision system. Our system consists of an event-based camera that captures driver movements as spike inputs and an optimized convolutional spiking neural network (CSNN) capable of efficient inference on 720 x 720 event streams without the need for computationally expensive preprocessing. The dataset includes 13 driver motion categories, classified by direction, lighting conditions, and participants, including challenging environments like low-light conditions and tunnels.

Our experiments show that the system achieved a high accuracy of 94.04\% in recognizing driver motions, even under difficult conditions such as varying light and directional inputs. This demonstrates that our proposed four-layer CSNN, designed for training and inference on energy- and resource-constrained platforms, can meet performance requirements. In addition, our experiments display the capability of distinguishing between unexpected abnormal driver motions and normal driver motions with an accuracy of 97.24\%. Our system was especially able to classify all the abnormal driver motions, indicating high reliability in sensing various abnormal driver motions when deployed in real applications. We also demonstrated that our system was comparable to other applications that implemented direct use of event vision systems with SNNs, such as the DVS Gesture recognition task, where our proposed system achieved similar accuracy with a less complex model structure. When deployed on Intel's Loihi 2 neuromorphic processor, the system demonstrated significant energy efficiency, with an energy-delay product (EDP) that was over 20,721 times more efficient than the RTX 3080 and 541 times more efficient than the Jetson Xavior NX. This level of efficiency is critical for real-time, on-device AI applications where power consumption is a key constraint.

By eliminating the need for time-consuming preprocessing and significantly reducing power consumption while maintaining high accuracy, our system can advance the safety and efficiency of autonomous driving systems using neuromorphic vision technology. Furthermore, the N-DriverMotion dataset provides a valuable resource for future research in driver behavior prediction, particularly in AI-driven systems requiring a high-resolution event dataset.

\section*{ACKNOWLEDGMENT}
This work was supported by the National Research Foundation of Korea (NRF) grant funded by the Korea government (MSIT) (NRF-) and the Ministry of Education (NRF-).

\bibliographystyle{unsrt}  


\end{document}